\documentclass[lettersize,journal]{IEEEtran}
\usepackage{amsmath,amsfonts}
\usepackage{algorithmic}
\usepackage{algorithm}
\usepackage{array}
\usepackage[caption=false,font=normalsize,labelfont=sf,textfont=sf]{subfig}
\usepackage{textcomp}
\usepackage{stfloats}
\usepackage{url}
\usepackage{verbatim}
\usepackage{graphicx}
\usepackage{multirow}
\usepackage{colortbl}
\usepackage[table,xcdraw]{xcolor}
\usepackage{threeparttable}
\usepackage{hhline}
\usepackage{threeparttable}
\usepackage{array}  % 引入array宏包以支持自定义列类型
\usepackage{tabularx}
\usepackage{booktabs}
\usepackage{hyperref}
\hypersetup{
	colorlinks=true,
	linkcolor=blue,
	anchorcolor=blue,
	citecolor=blue}

\begin{document}
\title{A Dual-stage Prompt-driven Privacy-preserving Paradigm for Person Re-Identification}
\author{Ruolin Li, Min Liu, Yuan Bian, Zhaoyang Li, Yuzhen Li, Xueping Wang and Yaonan Wang
	\thanks{This work was supported in part by the National Natural Science Foundation of China under Grant 62425305, U22B2050 and 62221002, in part by the Science and Technology Innovation Program of Hunan Province under Grant 2023RC1048, in part by the Hunan Provincial Natural Science Foundation of China under Grant 2024JJ3013. (\textit{Corresponding author: Min Liu.}) }
	\thanks{
		Ruolin Li, Min Liu, Yuan Bian, Zhaoyang Li, Yuzhen Li and Yaonan Wang are with the School of Artificial Intelligence and Robotics at Hunan University and National Engineering Research Center of Robot Visual Perception and Control Technology, Changsha, 410082, Hunan, China (e-mail: liruolin$@$hnu.edu.cn; liu\_min$@$hnu.edu.cn; yuanbian$@$hnu.edu.cn; zhaoyli$@$hnu.edu.cn; zzrs$@$hnu.edu.cn; yaonan$@$hnu.edu.cn).
		
		Xueping Wang is with the College of Information Science and Engineering at Hunan Normal University, Changsha, 410081, Hunan, China (e-mail: wang\_xueping$@$hnu.edu.cn).}}

%\markboth{Journal of \LaTeX\ Class Files,~Vol.~18, No.~9, September~2020}%
%{How to Use the IEEEtran \LaTeX \ Templates}

\maketitle

\begin{abstract}
With growing concerns over data privacy, researchers have started using virtual data as an alternative to sensitive real-world images for training person re-identification (Re-ID) models. However, existing virtual datasets produced by game engines still face challenges such as complex construction and poor domain generalization, making them difficult to apply in real scenarios. To address these challenges, we propose a Dual-stage Prompt-driven Privacy-preserving Paradigm (DPPP). In the first stage, we generate rich prompts incorporating multi-dimensional attributes such as pedestrian appearance, illumination, and viewpoint that drive the diffusion model to synthesize diverse data end-to-end, building a large-scale virtual dataset named GenePerson with 130,519 images of 6,641 identities. In the second stage, we propose a Prompt-driven Disentanglement Mechanism (PDM) to learn domain-invariant generalization features. With the aid of contrastive learning, we employ two textual inversion networks to map images into pseudo-words representing style and content, respectively, thereby constructing style-disentangled content prompts to guide the model in learning domain-invariant content features at the image level. Experiments demonstrate that models trained on GenePerson with PDM achieve state-of-the-art generalization performance, surpassing those on popular real and virtual Re-ID datasets.
\end{abstract}


\begin{figure}[t]
	\centering
	\includegraphics[width=1.0\columnwidth]{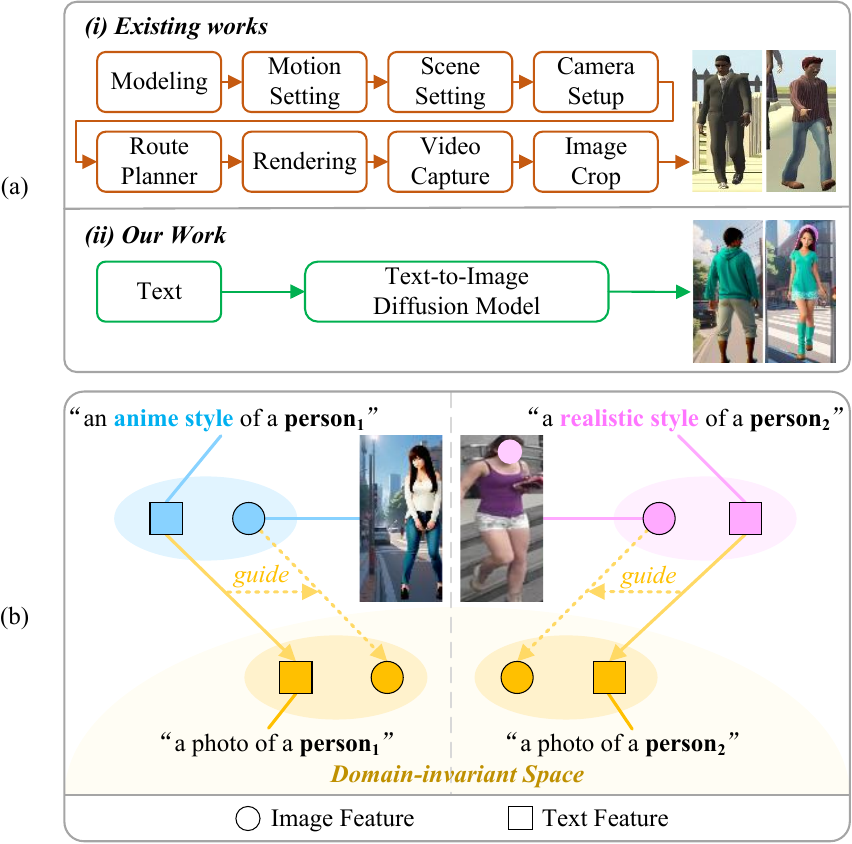}
	\caption{Motivation of our method. (a) Our method generates diverse virtual samples with only text, greatly simplifying the dataset construction process. (b) PDM learns domain-invariant visual content features under the guidance of prompts in the joint vision-language space.}
	\label{fig:1}
\end{figure}

\section{Introduction}
\IEEEPARstart{P}{erson} re-identification (Re-ID) aims to match the identity of the same person across camera views and is widely used in video surveillance, smart security, and other fields \cite{tan2025cameraagnostic,liu2025occuledAOANet,pang2024crossCHCR,peng2024MRLReID,liu2024by,ye2024SecureReID,tang2024NAS-PED,huang2022feature}. Most Re-ID methods rely on representation learning using real-world images captured by cameras \cite{gao2024semanticoccluded,Chen_2023_appearances,liu2024wf,pan2023posevideo,Yang_2024_promptsg,bianyuan2024}. However, collecting large-scale pedestrian datasets is not only costly, but also raises personal privacy concerns, as these raw images usually contain sensitive information such as identity information and personal whereabouts. Once a Re-ID model is illegally accessed, these data can be leaked and misused, posing serious threats to public safety. Due to these privacy issues, several Re-ID datasets \cite{dukemtmc-reid,msmt17} have been withdrawn in recent years, leading to a shortage of publicly available datasets and limiting further research of Re-ID.

Recently, researchers have increasingly focused on privacy-preserving Re-ID, aiming to protect data while maintaining model utility. Current approaches primarily include learnable anonymization \cite{ye2024SecureReID,Maximov2020CIAGAN,PI-Net2021Chen,Dietlmeier_2022_WACV,zhang2022learnableanony,kansal2024dualstageanony}, federated learning \cite{zhuang2020fedreid,Zhuang_2021fedureid,Wu_2021fedreid,ma2024fedsh}, and virtual data \cite{BARBOSA2018SOMAset,bar2018SyRI,sun2019PersonX,Xiang_2022_SynPerson,xiang2023FineGPR,wang2020RandPerson,Zhang_2021_UnrealPerson}. The first two methods still require the use of real data during training, which may lead to potential privacy risks such as information leakage or gradient inversion attacks \cite{zhu2019_dlg,zhao2020idlg}. In contrast, virtual data-based approaches fundamentally eliminate reliance on real-world images, offering stricter privacy protection constraints. However, existing virtual Re-ID datasets are mostly synthesized using game engines, which still face several challenges: 1) the construction of existing virtual datasets involves multiple steps, including clothing texturing, character modeling, motion setting, route planning, scene setup, rendering, data collection, and annotation, which is inherently complex and reliant on manual intervention; and 2) there exists a significant domain gap between virtual and real data, making it difficult for models to directly transfer features learned from virtual data to real-world scenarios, resulting in poor generalization performance on real datasets.
%character modeling, trajectory simulation, pedestrian localization, rendering, and data annotation

To address these challenges, we propose a Dual-stage Prompt-driven Privacy-preserving Paradigm, dubbed DPPP. DPPP aims to establish a low-cost virtual data generation pipeline and effectively mitigate the domain gap between virtual and real domains, thereby ensuring privacy preservation while maintaining the performance of the Re-ID model. In the first stage, we leverage a large language model to construct rich and fine-grained prompts that consider multiple attributes such as clothing, body shape, pose, and lighting, and introduce a conditional control module along with lightweight adaptation strategies to precisely control the diffusion model \cite{Rombach_2022_sd}, thus directly generating high-quality and diverse virtual data. As shown in Fig. \ref{fig:1}(a), our generation method employs a simplified one-step process that bypasses the complex multi-step procedures of traditional virtual dataset construction. In the second stage, we propose a Prompt-driven Disentanglement Mechanism (PDM), inspired by recent studies \cite{Bose_2024_stylip,Cho_2023_PromptStyler,kwon-2022-CLIPstyler,Fahes_2023_poda,gal2022textualinversion,Li_Zhang_Lan_Jiang_2025,zheng2025towards,zheng2024odtrack} showing that text representations can serve as prototypes for characterizing different image styles within the vision–language joint embedding space (i.e., the embedding space of CLIP \cite{pmlr-2021-clip}). As shown in Fig. \ref{fig:1}(b), each image can be represented in the joint vision-language space by a style-content prompt (“a [S] style of a [content]”). Due to the absence of semantic labels in most Re-ID datasets and the diversity and variability of pedestrian image styles, it is challenging to accurately describe the content and style of pedestrian images with concrete words. Consequently, we employ two separate textual inversion networks to learn style and content pseudo-words that represent specific visual contexts for more flexible image semantic modeling. The learned content pseudo-word is subsequently used to construct a style-disentangled content prompt (“a photo of a [content]”) as a prototype, guiding the model to learn domain-invariant visual content features, thereby mitigating the impact of style variations on its generalization to real-world scenarios. We align the text representations of style-content prompts and content prompts with their corresponding image representations through contrastive learning, enabling effective modeling of style and content.

Our contributions are summarized as follows: 
\begin{itemize}
	\item We propose DPPP, a novel dual-stage prompt-driven virtual data privacy-preserving paradigm that utilizes language prompts to control virtual image generation and guide the model in learning domain-invariant features.
	\item To the best of our knowledge, this work is the first to apply a diffusion model to privacy-preserving Re-ID, enabling one-step automatic generation of virtual data. Moreover, a large-scale generative Re-ID dataset named GenePerson is constructed, comprising 130,519 images of 6,641 identities in a variety of outfits, poses, and scenes to ensure diversity.
	\item A novel disentanglement mechanism based on the joint vision-language space is proposed, which independently models image style and content through text, and further utilizes content-relevant text features to guide the learning of domain-invariant visual content representations, thus enhancing generalization to real-world domains.
	\item Experimental results demonstrate that models trained on GenePerson exhibit better generalization than those trained on other widely used real-world and virtual Re-ID datasets, with the proposed PDM further improving the results.
\end{itemize}

\section{Related Works}

\subsection{Privacy Preservation in Re-ID}
  Efforts to address data privacy concerns can be grouped into three categories: learnable anonymization \cite{ye2024SecureReID,Maximov2020CIAGAN,PI-Net2021Chen,Dietlmeier_2022_WACV,zhang2022learnableanony,kansal2024dualstageanony}, federated learning \cite{zhuang2020fedreid,Zhuang_2021fedureid,Wu_2021fedreid}, and virtual data \cite{BARBOSA2018SOMAset,bar2018SyRI,sun2019PersonX,wang2020RandPerson,Zhang_2021_UnrealPerson,Xiang_2022_SynPerson,xiang2023FineGPR}. The first two approaches assume that visual information about the target is available. However, the collection of datasets has become extremely difficult due to increasingly stringent privacy restrictions. For example, DukeMTMC-reID \cite{Zheng_2017_DukeMTMC} was withdrawn due to privacy concerns; meanwhile, some European countries have enacted laws prohibiting unauthorized collection and use of personal data \cite{Michelle2017GDPR,Bhaimia_2018}. In this context, virtual data-based methods are gradually gaining attention. Virtual data-based methods can effectively re-identify real-world individuals using only synthetic virtual data training, which fundamentally eliminates the privacy issues that may arise from training with real images. Barbosa \textit{et al.} \cite{BARBOSA2018SOMAset} manually create SOMAset, which consists of 50 human models. Bak \textit{et al.} \cite{bar2018SyRI} propose SyRI, which renders 100 virtual identities under different lighting conditions using HDR environment maps. Sun \textit{et al.} \cite{sun2019PersonX} propose PersonX, a dataset comprising 1,266 characters with 36 different viewpoints. SynPerson \cite{Xiang_2022_SynPerson} and FineGPR \cite{xiang2023FineGPR} datasets both consider weather conditions. Wang \textit{et al.} \cite{wang2020RandPerson} propose a large-scale synthetic dataset RandPerson containing 8,000 characters. Zhang \textit{et al.} \cite{Zhang_2021_UnrealPerson} propose another large synthetic dataset UnrealPerson, which contains 3,000 characters and renders scenes with more realism.

  Essentially, these game engine-based virtual data methods rely on complex multi-step production pipelines and multiple software tools. In contrast, our approach achieves end-to-end generation through a diffusion model \cite{Rombach_2022_sd} that can directly synthesize a variety of virtual pedestrian images based on text prompts we generate. This text-to-image generation paradigm not only simplifies the dataset construction pipeline, but also provides the flexibility to create various samples through natural language.
  
\subsection{Domain Generalization}
 A major challenge of virtual data lies in the lack of real images and the large domain shift between virtual and real data, resulting in limited generalization of trained models in real-world deployments. To mitigate such domain gap, Domain Generalization (DG) aims to learn domain-invariant representations that can generalize to unseen target domains. Traditional DG methods rely on models trained exclusively on image data \cite{Zhou2020aaai-dg,zhou2021MixStyle,Zhou2020eccv-dg,kang2022sytledg}. Recently, with the advancement of large-scale vision-language models (i.e., CLIP \cite{pmlr-2021-clip}), several DG methods have begun exploring integration with prompt engineering to better leverage semantic information for domain generalization \cite{Bose_2024_stylip,Cho_2023_PromptStyler,yu_2024_CLIPCEIL,vidit2023clipthegap,cha2022MIRO}. These large-scale vision-language models are trained on massive image-text pairs, aligning image and text representations in a shared semantic space. Various visual tasks focusing on generalization can utilize this space to efficiently manipulate visual features through domain-invariant prompts \cite{kwon-2022-CLIPstyler,Fahes_2023_poda,Patashnik-2021-StyleCLIP,Dunlap_2023_LADS,gal-2022-StyleGAN-NADA,yang2024PDCL-Attack,zhang2024badcm,wang2024Cross-ModalRetrieval}. Yang \textit{et al.} \cite{yang2024PDCL-Attack} employs CLIP as a tool for text-driven feature manipulation and domain expansion. Fahes \textit{et al.} \cite{Fahes_2023_poda} leverages textual prompts from the target domain to guide the mapping of source features in the CLIP latent space. Cho \textit{et al.} \cite{Cho_2023_PromptStyler} models various image style distributions in the CLIP latent space through textual prompts to enhance the model’s domain generalization ability. These works rely on semantic labels to construct textual descriptions, which are unavailable in Re-ID where only index-based identity labels are provided. To address this limitation, we introduce textual inversion \cite{gal2022textualinversion} to encapsulate visual context into prompts, serving as an alternative to semantic annotation. Also, unlike existing work \cite{Yang_2024_promptsg} that inverts an image into a rough global pseudo-word token, we decompose the inversion task into two dimensions, content and style.

 \begin{figure*}[t]
 	\centering
 	\includegraphics[width=2.0\columnwidth]{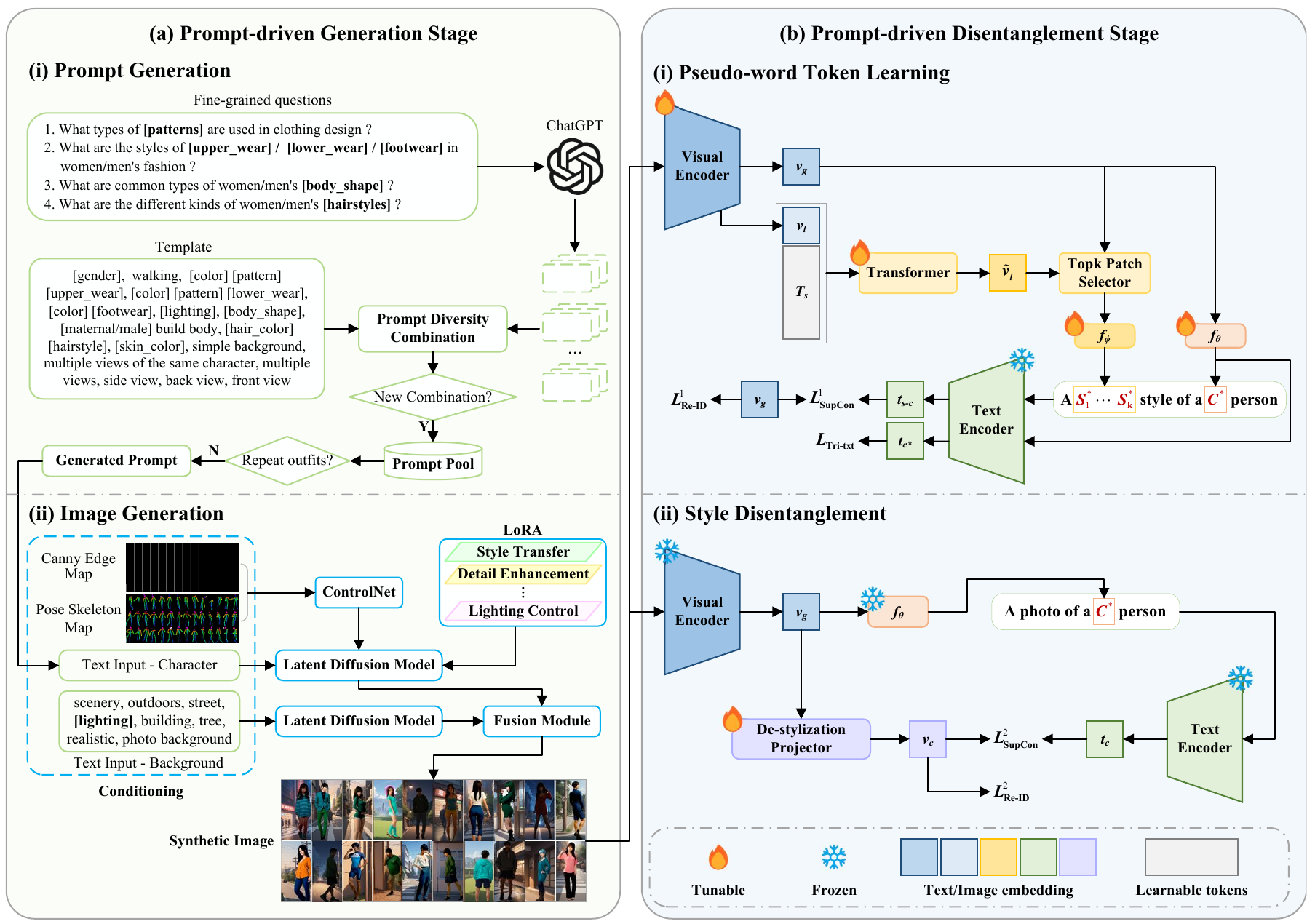}
 	\caption{Overview of our framework, which consists of two parts. (a) The prompt-driven virtual image generation pipeline shows the generation of our dataset GenePerson. (b) The disentanglement stage first uses text to effectively capture the style and content information of the image, and then utilizes style-disentangled prompts as a guide for disentangling the visual representations.}
 	\label{fig:2}
 \end{figure*}
 
\section{Methodology}

\subsection{Problem Definition}
We consider a privacy-preserving Re-ID task , where access to the appearance of real pedestrians is restricted. Therefore, we construct a large-scale virtual person dataset via a text-to-image diffusion model, which serves as the source domain, denoted by $D^s=\{x_i^s,y_i^s\}^{n^s}_{i=1}\approx{P^s}$, where $x_i^s\in\mathcal{X}^s$, $y_i^s\in\mathcal{Y}^s$, and $P^s$ represent the input data, identity label, and the joint distribution of the data and the labeling space, respectively. The goal is to train a model on $D^s$ that can be directly implemented to an unseen real-world target domain $D^t=\{x_i^t,y_i^t\}^{n^t}_{i=1}\approx{P^t}$, where $x_i^t\in\mathcal{X}^t$ and $y_i^t\in\mathcal{Y}^t$ and $P^t$ denotes the target distribution. In our work, we consider $P^t\neq{P^s}$, indicating that the target and source distribution are mutually distinct.

\begin{figure}[t]
	\centering
	\includegraphics[width=1.0\columnwidth]{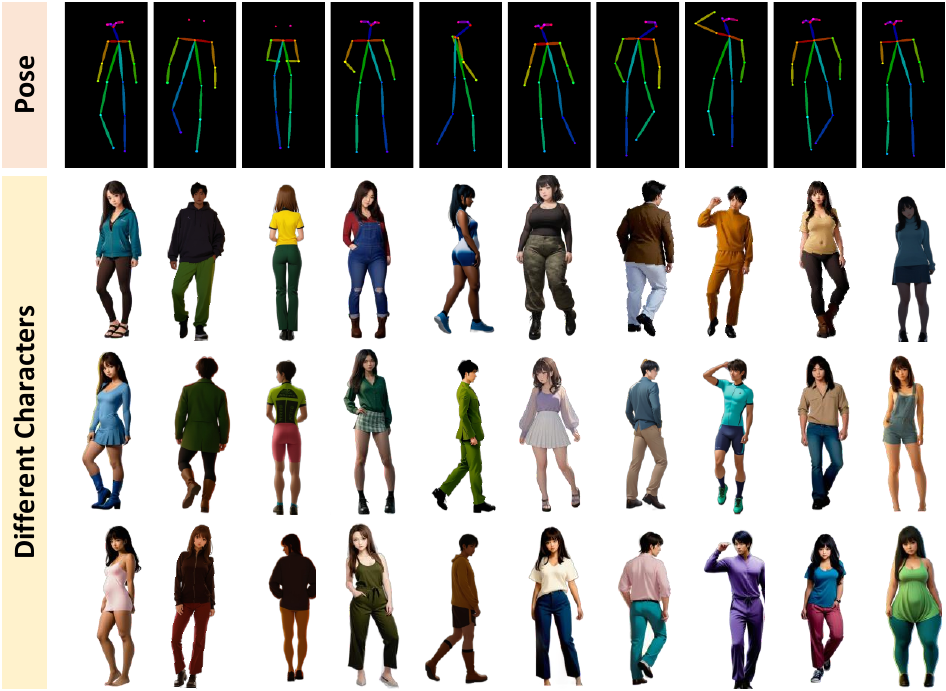}
	\caption{Pedestrians with different postures in the GenePerson dataset. Generate corresponding virtual samples based on the given poses.}
	\label{fig:3}
\end{figure}

\begin{figure}[t]
	\centering
	\includegraphics[width=1.0\columnwidth]{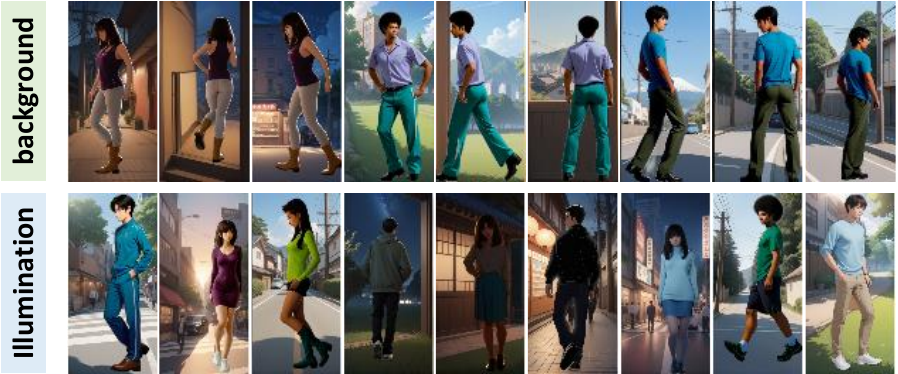}
	\caption{Some of the image samples in the proposed GenePerson dataset, including 3 different pedestrians in different background scenes, and pedestrians in 9 different illumination conditions.}
	\label{fig:4}
\end{figure}

\subsection{Image Generation Pipeline}
In the first stage, we propose a prompt-driven virtual data auto-generation pipeline for building our dataset GenePerson. As shown in Fig. \ref{fig:2}(a), the overall production process contains two core modules, prompt generation and image synthesis.

\textbf{Character Prompt Generation.} Text-to-image synthesis models rely on semantic control from the input of textual prompts. As shown in Fig. \ref{fig:2}(a)(i), we design a prompt template containing multiple placeholders, each representing key visual attributes of a character, such as clothing style, body shape, hairstyle, and skin color. By substituting these placeholders with specific terms in different combinations, we can generate rich prompts. We utilize ChatGPT to generate candidate attribute descriptions through the following questions: 1) “Please list specific [color] names”; 2) “What types of [patterns] are used in clothing design”; 3) “What are the styles of clothing”; 4) “What are the different types of body shapes”; 5) “What are the different kinds of hairstyles”, along with additional questions such as “Please list specific [color] names” and “Please list the common skin colors”. The responses are then inserted into the corresponding placeholders to form initial textual prompts. These initial prompts are subsequently randomly combined based on the attributes to generate diverse descriptions, with deduplication applied to the combinations of upper wear, lower wear, and footwear color and style, as clothing is a crucial visual cue for identity recognition. For example, a final prompt might read: “girl, walking, pink plaid shirt, black distressed jeans, white loafers, day, sunshine, slim figure, maternal build body, dark brown hair color, waist-length hair, straight hair texture, high ponytail, bangs, warm beige skin, simple background, multiple views of the same character, multiple views, side view, back view, front view”.

\textbf{Character Generation.} We introduce the Stable Diffusion Model (SDM) \cite{Rombach_2022_sd} to end-to-end synthesize diverse characters based on gender, body shape, styling, viewpoint, illumination, and other factors via generated prompts. Although the native SDM excels in generic images, it remains limited in rendering details such as clothing textures, poses, limbs, lights and lighting, making it insufficient for fine-grained Re-ID task \cite{zhu2025tonemap}. To generate reasonable characters, we explore LoRA \cite{hu2022lora} for targeted enhancements: 1) an Image Enhancement LoRA to improve the control of the model over visual details; 2) a Lighting Adaptation LoRA to enhance adaptability to complex lighting scenarios; and 3) a Style Modulation LoRA to achieve more natural appearances in character generation. Further, we integrate ControlNet \cite{zhang2023controlnet} for finer control through additional inputs from conditional images. We input an edge map along with multiple pose skeleton maps to simultaneously generate multi-pose images of the same pedestrian. Fig. \ref{fig:3} illustrates the characters generated based on the specified poses. Ultimately, we generate 6,641 virtual pedestrians.

\textbf{Scenario Design.} To construct a scene-diverse dataset, we employ SDM to generate background images of indoor and outdoor scenes, such as streets, commercial areas, parks, fields and residential areas. We embed different simple temporal lighting descriptions in the background prompts, such as “day, sunshine”, “dusk” and “night, dark”. Finally, we perform pixel-level fusion of the segmented virtual characters with the generated background images according to the lighting conditions. Fig. \ref{fig:4} shows images of the same pedestrian with different backgrounds, and the pedestrian images under different lighting conditions in the GenePerson dataset.

\textbf{Data Annotation.} Our method simultaneously generates multi-view images of the same identity on a single canvas, enabling automatic cropping and labeling without additional data collection, thus achieving zero-cost annotation.

\subsection{Pseudo-word Token Learning}
As shown in Fig. \ref{fig:2}(b)(i), to capture the style and content information of each input virtual image, we utilize two independently parameterized textual inversion networks to learn trainable tokens corresponding to the style and content pseudo-words in the style-content prompt “a $[S_1^*, \cdots, S_k^*]$ style of a $[C^*]$ person”. Note that the text encoder is kept frozen during this learning stage.

\textbf{Content Pseudo-word Token.} We aim to encapsulate the domain-shared identity information in an image into a content pseudo-word $C^*$. Specifically, we extract the global visual embedding $v_g\in\mathbb{R}^{d_1}$ from the final layer of the CLIP visual encoder, where $d_1$ denotes the dimension of the global feature. A multi-layer perceptron (MLP) is employed as an inversion network to invert $v_g$ into a pseudo-word token $c^*$. This process can be formalized as:
%, capturing the overall features of the image
\begin{equation}
	c^* = f_\theta \left( v_g \right),
	\label{eq:eq1}
\end{equation}
where $f_\theta \left( \cdot \right)$ denotes an inversion network parameterized by $\theta$, and $c^* \in T^*$, where $T^*$ denotes the token embedding space.

\textbf{Style Pseudo-word Token.} Style is usually expressed as a combination of local visual elements such as color distribution and texture structure, which relies on the combination of multiple related local regions rather than by a single image patch. To effectively model image style, we assume each image contains $n$ potential style elements, represented by a set of learnable tokens corresponding to style pseudo-words $S^*=\{S_i^*\}^k_{i=1}$, which are learned through adaptive aggregation of local patch features. Additionally, considering that the deepest layers of visual encoders primarily capture global semantic information such as identity or category while lacking local pattern details \cite{lin2024multi,zhu2024weather,zheng2016good}, we extract local features from the penultimate layer as it better preserves low and mid-level visual attributes related to image style.
	
Let $V=\{v_l^i\}_{i=1}^m\in\mathbb{R}^{d_2\times{m}}$ denote the patch features output from the penultimate layer of encoder, where $m$ is the number of image patches, and $d_2$ is the feature dimension for each patch. We define a set of $n$ learnable tokens $T_s=\{T_i\}^n_{i=1}\in\mathbb{R}^{d_2\times{n}}$, where each token corresponds to a latent style element. These tokens and $V$ are jointly fed into a Transformer block for adaptive optimization. To ensure alignment with CLIP’s textual embedding space, a fully connected layer is utilized to project the output into the target dimension. The entire process can be formalized as:
%Therefore, we choose the penultimate layer of the visual encoder to extract local features, as it better preserves medium and low level visual patterns associated with style while mitigating entanglement with high-level semantics.
\begin{equation}
	\tilde{V} = \mathrm{FC} \left( \mathrm{Transformer} \left( \left[ T_s | V \right] \right) \right),
	\label{eq:eq2}
\end{equation}
where $\tilde{V}=\{\tilde{v}_l^i\}^n_{i=1}\in\mathbb{R}^{d_1\times{n}}$ denotes the style features of the image, and $\left[\cdot|\cdot \right]$ means concatenation.

Notably, our approach strictly aligns the discriminative dimension during the style disentanglement, which is expected to ensure that the learned style features focus only on visual features relevant to identity matching. To mitigate interference from irrelevant noisy style representations, we introduce a Global Feature-based Local Filtering Module (GFLFM). GFLFM utilizes global image features as a soft reference to evaluate the relevance of each style feature, and retains only those with high relevance. Firstly, for the i-th style feature $\tilde{v}_l^i$, its attention weight $w_i$ with the global feature $v_g$ is computed as:
\begin{equation}
	w_i = \frac{\exp \left( \left( \tilde{v}_l^i \right)^\top v_g \right)}{\sum_{j=1}^{n} \exp \left( \left(\tilde{v}_l^j \right)^\top v_g \right)}.
	\label{eq:eq3}
\end{equation}
Then, based on the computed weights $W=\{w_i\}^n_{i=1}$, we sort the style features and select the top $k$ ones with the highest scores:
\begin{equation}
	V_L = \mathrm{TopK} \left(\tilde{V}, W, k \right),
	\label{eq:eq4}
\end{equation}
 where $V_L\in\mathbb{R}^{d_1\times{k}}$ means the final selected style features.
 
 Similarly, we employ an MLP-based inversion network, parameterized by $\phi$, to invert the selected style features into a set of pseudo-word tokens. This can be mathematically described as:
 \begin{equation}
 	\left[s^*_1,\cdots,s^*_k \right] = f_\phi \left(V_L \right).
 	\label{eq:eq5}
 \end{equation}

\textbf{Cross-modal Contrastive Learning.} After obtaining the content and style pseudo-word tokens, each image can be represented as a structured textual description, which is then fed into the frozen CLIP text encoder to obtain the corresponding text feature, denoted as $t_{s\text{-}c}$. To encourage pseudo-words to efficiently encapsulate corresponding visual contexts belonging to the same identity, we apply the following symmetric supervised contrastive loss:
\begin{equation}
	\mathcal{L}_{\text{SupCon}}^1 = \mathcal{L}_{\text{i2t}} + \mathcal{L}_{\text{t2i}},
	\label{eq:eq6}
\end{equation}
\begin{equation}
	\mathcal{L}_{\text{i2t}} = \frac{1}{B} \sum_{i=1}^{B} \sum_{p^+ \in P(j))}
	\log \frac{\exp \left( \text{sim} \left( v_g^i,t_{s\text{-}c}^{p^+} \right)/\tau \right)}
	{\sum_{j=1}^{B} \exp \left( \text{sim} \left( v_g^i,t_{s\text{-}c}^{j} \right)/\tau \right)},
	\label{eq:eq7}
\end{equation}
\begin{equation}
	\mathcal{L}_{\text{t2i}} = \frac{1}{B}\sum_{i=1}^{B}\sum_{p^+\in{P(j))}}
	\log \frac{\exp \left( \text{sim} \left( t_{s\text{-}c}^i,v_g^{p^+} \right)/\tau \right)}
	{\sum_{j=1}^{B} \exp \left( \text{sim} \left( t_{s\text{-}c}^i,v_g^{j} \right)/\tau \right)},
	\label{eq:eq8}
\end{equation}
where $v_g^i$ and $t_{s\text{-}c}^i$ denote the global image feature and the style-content text feature of the $i$-th image in a batch of size $B$, respectively. $P(j)$ denotes the positive samples associated with $v_g^i$ and $t_{s\text{-}c}^i$. $\tau$ is a temperature hyperparameter.

To enhance the identity recognition of content pseudo-words, we utilize a triplet loss to encourage pseudo-words with the same identity to cluster in the text embedding space, while pushing apart those with different identities. The content pseudo-word tokens are fed into the CLIP text encoder to obtain their corresponding text representations $t_{c^*}$. Given an anchor $t_{c^*}^a$, a positive $t_{c^*}^p$, and a negative $t_{c^*}^n$, the triplet loss is formulated as:
\begin{equation}
 	\mathcal{L}_{\text{Tri-txt}} = \max\left( \|t_{c^*}^a - t_{c^*}^p\|_2^2 - \|t_{c^*}^a - t_{c^*}^p\|_2^2 + \delta, 0 \right),
	\label{eq:eq9}
\end{equation}
where $\|\cdot\|_2$ denotes the Euclidean distance, and $\delta$ is a margin hyperparameter.

We also employ the standard Re-ID loss, i.e., the triplet loss and the identity loss \cite{he2023fastreid}, to optimize the image encoder:
\begin{equation}
	\mathcal{L}_{\text{Re-ID}}^1 = \mathcal{L}_{\text{Tri-img}}^1+\mathcal{L}_{\text{ID}}^1,
	\label{eq:eq10}
\end{equation}
\begin{equation}
	\mathcal{L}_{\text{Tri-img}}^1 = \text{max}\left(d_p-d_n+\alpha \right),
	\label{eq:eq11}
\end{equation}
\begin{equation}
	\mathcal{L}_{\text{ID}}^1 = -\frac{1}{B}\sum_{j=1}^B\text{log}p_j,
	\label{eq:eq12}
\end{equation}
where $p_j$ is the ID prediction probability for the $j$-th class, $d_p$ and $d_n$ represent the feature distances of the positive and negative pairs, and $\alpha$ denotes the margin. Notably, in accordance with CLIP-ReID \cite{Li_2023_clipreid}, we use the features before and after the linear layer following the transformer for the computation here, and additionally compute $\mathcal{L}_{\text{Tri-img}}^1$ after the 11-th transformer layer.

The overall function is defined as follows:
\begin{equation}
	\mathcal{L}_{\text{PTL}} = \mathcal{L}_{\text{Re-ID}}^1+\mathcal{L}_{\text{SupCon}}^1+\mathcal{L}_{\text{Tri-txt}}.
	\label{eq:eq13}
\end{equation}

\subsection{Prompt-driven Style Disentanglement}
After learning the style and content pseudo-words, as depicted in Fig. \ref{fig:2}(b)(ii), we further perform the style disentanglement at the image level guided by the content text features. Concretely, we design a de-stylization projector $P_s$ to filter out the style information from the raw image features, obtaining the visual content features, denoted as $v_c\in\mathbb{R}^{d_1}$. The process is defined as:
\begin{equation}
	v_c = P_s(v_g).
	\label{eq:eq14}
\end{equation}
Be aware that during this learning stage, only the parameters of $P_s$ are tunable. 

With the aim of learning visually domain-invariant representations, we encourage $v_c$ to gather around style-disentangled content prompts. Specifically, the trained content inversion network $f_\theta$ is kept fixed, and a content prompt is constructed as: “a photo of a $[C^*]$ person”, where $C^*$ is the learned content pseudo-word. This content prompt is input to a frozen CLIP text encoder to derive content-oriented text features $t_c$, while $v_c$ are encouraged to align with $t_c$ through symmetric supervised contrastive loss, thereby enabling the learning of visually modality-shared identity content. The loss is formulated as:
\begin{equation}
	\begin{split}
		\mathcal{L}_{\text{SupCon}}^2 &= \frac{1}{B} \sum_{i=1}^{B} \sum_{p^+ \in P(j)} \Bigg( 
		\log \frac{\exp \left( \text{sim} \left( v_c^i, t_c^{p^+} \right)/\tau \right)}
		{\sum_{j=1}^{B} \exp \left( \text{sim} \left( v_c^i, t_c^j \right)/\tau \right)} \\
		&+ \log \frac{\exp \left( \text{sim} \left( t_c^i, v_c^{p^+} \right)/\tau \right)}
		{\sum_{j=1}^{B} \exp \left( \text{sim} \left( t_c^i, v_c^j \right)/\tau \right)} 
		\Bigg).
	\end{split}
	\label{eq:eq15}
\end{equation}
We perform Re-ID loss computation using $v_c$ to enhance the modeling of identity-related features. Ultimately, the overall loss for training $P_s$ is defined as:
\begin{equation}
	\mathcal{L}_{\text{PSD}} = \mathcal{L}_{\text{Re-ID}}^2 + \mathcal{L}_{\text{SupCon}}^2.
	\label{eq:eq16}
\end{equation}

\begin{figure}[t]
	\centering
	\includegraphics[width=1.0\columnwidth]{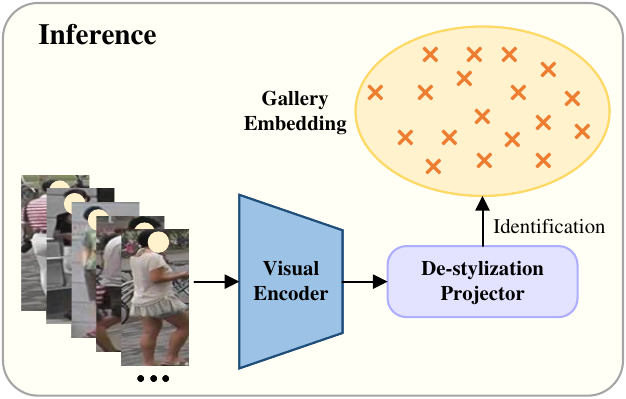}
	\caption{At inference time, the trained visual encoder and de-stylization projector are used to extract image content features.}
	\label{fig:5}
\end{figure}

\subsection{Inference}
As shown in Fig. \ref{fig:5}, we employ the trained visual encoder and de-stylization projector for inference. Given a query image $x^t\in{D^r}$ from the real domain, the visual encoder extracts its original visual feature $v_r\in\mathbb{R}^{d_1}$. Subsequently, $v_r$ is passed through the de-stylization projector to obtain the visual content feature $\tilde{v_r}$. We take $\tilde{v_r}$ as the final identity representation, which is matched for similarity with the gallery corresponding embedding.

\begin{table}[t]
	\centering
	\caption{The statistics of real-world datasets in our experiments.}
	\normalsize
	\renewcommand\arraystretch{1.2}
	\setlength{\tabcolsep}{0pt}
	\begin{threeparttable} 
		\begin{tabular*}{\columnwidth}{@{\extracolsep{\fill}}
				>{\centering\arraybackslash}p{4cm}|
				>{\centering\arraybackslash}p{1.5cm}|
				>{\centering\arraybackslash}p{1.5cm}|
				>{\centering\arraybackslash}p{1.5cm}
			}
			\hline
			Dataset & \#ID & \#Images & \#Cams \\
			\hline\hline
			Market-1501 \cite{zheng2015market1501} & 1,501 & 32,668 & 6 \\
			DukeMTMC-reID \cite{dukemtmc-reid} & 1,404 & 36,411 & 8 \\
			\hline
		\end{tabular*}
	\end{threeparttable}
	\label{tab:1}
\end{table}

\begin{table}[t]
	\centering
	\caption{A comparison of some synthetic datasets. “Generation” refers to data generation techniques.}
	\normalsize
	\renewcommand\arraystretch{1.2}
	\setlength{\tabcolsep}{1pt}
	\begin{threeparttable}
		\begin{tabular*}{\columnwidth}{@{\extracolsep{\fill}}
				>{\centering\arraybackslash}p{3cm}|
				>{\centering\arraybackslash}p{1.2cm}|
				>{\centering\arraybackslash}p{1.5cm}|
				>{\centering\arraybackslash}p{3cm}
			}
			\hline
			Dataset & \#ID & \#Images &  Generation \\
			\hline\hline
			SOMAset \cite{BARBOSA2018SOMAset} & 50 & 100,000  & Game Engine \\
			SyRI \cite{bar2018SyRI} & 100 & 1,680,000  & Game Engine \\
			PersonX \cite{sun2019PersonX} & 1,266 & 273,456  & Game Engine \\
			RandPerson \cite{wang2020RandPerson} & 8,000 & 132,145 & Game Engine \\
			UnrealPerson \cite{Zhang_2021_UnrealPerson} & 3,000 & 120,000 & Game Engine \\
			FineGPR \cite{xiang2023FineGPR} & 1,150 & 2,028,600 & Game Engine \\
			\hline
			GenePerson (Ours) & 6,641 & 130,519 & Generative Model \\
			\hline
		\end{tabular*}
	\end{threeparttable}
	\label{tab:2}
\end{table}

\section{Experiments}
\subsection{Experimental Settings}
	\textbf{Datasets and Evaluation Protocols.} In this paper, two widely used real-world person Re-ID datasets are used for generalization evaluation, including Market-1501 \cite{zheng2015market1501} and DukeMTMC-reID \cite{dukemtmc-reid}. The statistics of the real datasets are listed in Tab. \ref{tab:1}, and six existing synthetic datasets for comparison are described in Tab. \ref{tab:2}.
	%The Market-1501 dataset collects 32,668 images of 1,501 identities, shot by 6 cameras. Its training set contains 12,936 images of 751 pedestrians and the test set contains 19,732 images of the remaining 750 pedestrians. The CUHK03 dataset is repartitioned according to the CUHK03-NP protocol \cite{zhong2017cuhk03-np}, where 7,365 images of 767 identities are categorized as the training set, and the rest of the 6,732 images of 700 identities are used as the test set. The experiment uses detected bounding boxes instead of labeled ones. The DukeMTMC-reID dataset collects 36,411 images from 1,812 pedestrians recorded by 8 cameras. The training set contains 702 identities with 16,522 images and the other 1110 identities with 19,889 images are used for testing. The MSMT17 dataset, being a sizable pedestrian dataset, contains 126,441 images for 4,101 identities captured by 15 cameras. Among them, 32,621 images of 1,041 identities are used as training set and 93,820 images of 3,060 identities are used as testing set. Moreover, we selected six existing synthetic datasets for comparison, which are described in Tab. \ref{tab:1}.
	%Moreover, three existing synthetic datasets Randperson \cite{wang2020RandPerson}, PersonX \cite{sun2019PersonX}, and SyRI \cite{bar2018SyRI} are taken for comparison, which are described in the related work section.

	Following common practice in the Re-ID community, this work evaluates the performance using mean Average Precision (mAP) and Cumulative Matching Characteristics (CMC) at Rank-1 and Rank-5.

	\textbf{Implementation Details.} For the image generation stage, we select Stable Diffusion XL as the generative model and Euler a as the sampling algorithm. Human pose skeletons are pre-generated using OpenPose \cite{zhe2021openpose} and fed into ControlNet. Meanwhile, we prepare a gird map with 42 evenly distributed blocks, which is processed using the Canny edge detection algorithm \cite{canny1986} before being fed into ControlNet. Subsequently, each generated image is automatically divided into 42 equally-sized regions, which are assigned identity labels corresponding to the image. Finally, U2Net \cite{qin2020u2net} is used to segment the foreground person from each region, which is fused to the background based on the alpha channel.
	%For the image generation stage, we supplement its implementation details. First, we choose Stable Diffusion XL and Euler a sampling algorithm. Second, we draw on the Market1501 dataset as a real image reference, from which we pre-extract human pose skeletons using OpenPose \cite{zhe2021openpose} offline in order to introduce privacy-irrelevant pose information. These skeleton maps are cyclically fed into ControlNet as additional conditions to generate virtual person with one-to-one matching. Meanwhile, we prepare a grid map consisting of 42 evenly spaced patches, which is processed by the Canny edge detection algorithm \cite{canny1986} and fed into ControlNet to guide Stable Diffusion to generate the same character in 42 different viewpoints and poses on the same canvas simultaneously. Then, we automatically segmented the canvas into 42 separate images and assigned corresponding identity labels on a canvas basis. Eventually, we fuse the virtual characters with randomly selected synthetic background images according to the alpha channel to generate the final virtual samples. To be emphasized, the generated prompts for each background image contain specific lighting conditions, and this information is likewise incorporated into the prompts when generating the virtual characters. To ensure semantic consistency, we group virtual characters with backgrounds having corresponding lighting conditions for fusion based on these key descriptions.
	%To ensure data quality, we manually eliminate undesirable samples.

	For the feature disentanglement stage, we adopt the ViT-B/16 pre-trained CLIP as the backbone to extract features. Our framework adds randomly initialized inversion networks, Transformer and de-stylization projector. Both content and style inversion networks are designed as three-layer MLPs with hidden dimensions of 512 and 2048, respectively. The Transformer for extracting local style features of the image is set to 3 layers and 1 header, and the number of potential style elements $n$ is set to 24. Inspired by the projection head design in CLIP, the de-stylization projector is implemented as a lightweight two-layered perceptron. A batch normalization layer followed by a linear fully connected layer is placed at the end of the network. The number of effective local features $k$ in Eq. (4) is set to 6. The batch size is set to 64, containing 8 identities with 8 images per identity. All input images are resized to $256\times{128}$. Besides, to enhance intra-class variation, we apply image-level augmentation by independently sampling contrast coefficients in the range $(0.5, 1.5)$ for each image. We use the Adam optimizer for training, with an initial learning rate of 5e-6 for the visual encoder and 5e-5 for the randomly initialized modules. Both the visual encoder and the de-stylization projector are trained for 20 epochs, and the learning rate decay factor is reduced by a factor of 0.1 at the 15th and 20th epochs.
	
	The entire framework runs on a single NVIDIA RTX A6000 GPU with 48GB VRAM, where the training stage is implemented using PyTorch.

\begin{table*}[t]
	\centering
	\normalsize
	\caption{Performance Comparison with Existing Real and Synthetic Datasets on Market-1501 and DukeMTMC-reID, Respectively.}
	\renewcommand\arraystretch{1.2}
	\setlength{\tabcolsep}{0pt}
	\begin{threeparttable}
			\begin{tabular*}{\textwidth}{@{\extracolsep{\fill}}>{\centering\arraybackslash}p{4.5cm}|>{\centering\arraybackslash}p{1.5cm}|*{3}{>{\centering\arraybackslash}p{2cm}}|*{3}{>{\centering\arraybackslash}p{2cm}}}
				\hline
				\multicolumn{2}{c|}{Testing Set $\rightarrow$} & \multicolumn{3}{c|}{Market-1501} & \multicolumn{3}{c}{DukeMTMC-reID} \\
				\hline
				Training Set $\downarrow$ & Synthetic & Rank-1 & Rank-5 & mAP  & Rank-1 & Rank-5 & mAP \\
				\hline\hline
				Market-1501 \cite{zheng2015market1501} & $\times$     & -    &  -  & - & 30.7  &  45.0  & 15.0 \\
				DukeMTMC-reID \cite{dukemtmc-reid}   & $\times$     & 49.8 & 66.8 & 22.5  & -  & - & - \\
				SOMAset \cite{BARBOSA2018SOMAset}    & $\checkmark$ & 4.5 & - & 1.3  & 4.0  & -  & 1.0 \\
				SyRI \cite{bar2018SyRI}   & $\checkmark$ & 29.0 & - & 10.8  & 23.7  & - & 9.0 \\
				PersonX \cite{sun2019PersonX}    & $\checkmark$ & 44.0 & - & 20.4  & 35.4  & -  & 18.1 \\
				FineGPR \cite{xiang2023FineGPR}    & $\checkmark$ & 50.5 & 67.7 & 24.6  & -  & - & - \\
				RandPerson \cite{wang2020RandPerson}  & $\checkmark$ & 55.6 & - & 28.8  & 47.6 & - & 27.1 \\
				UnrealPerson \cite{Zhang_2021_UnrealPerson} & $\checkmark$ & 54.4 & 70.2 & 27.9  & 48.2 & 64.5 & 26.3 \\
				\hline\hline
				GenePerson (Ours) & $\checkmark$ & \textcolor{blue}{\textbf{57.0}} & \textcolor{blue}{\textbf{71.9}} & \textcolor{blue}{\textbf{31.9}} & \textcolor{blue}{\textbf{56.1}} & \textcolor{blue}{\textbf{70.5}} & \textcolor{blue}{\textbf{36.1}} \\					
				GenePerson$^{\dag}$ (Ours) & $\checkmark$ & \textcolor{red}{\textbf{57.7}} & \textcolor{red}{\textbf{73.0}} & \textcolor{red}{\textbf{32.6}} & \textcolor{red}{\textbf{57.5}} & \textcolor{red}{\textbf{71.3}} &  \textcolor{red}{\textbf{37.2}} \\
				\hline
			\end{tabular*}
		\begin{tablenotes}[flushleft]
			\item \textcolor{red}{\textbf{Red}} indicates the best and \textcolor{blue}{\textbf{blue}} the second best. A dagger ($^{\dag}$) means training on our PDM method. Unrealperson is extracted from unreal\_v1.1, unreal\_v2.1, unreal\_v3.1 and unreal\_v4.1.
		\end{tablenotes}
	\end{threeparttable}
	\label{tab:3}
\end{table*}
%An asterisk ($^*$) means that the results are reported from RandPerson \cite{wang2020RandPerson}.

\subsection{Comparison with State-of-the-art Methods}
We evaluate the generalization of GenePerson using direct transfer, which means training a model on a specific dataset and then evaluating its performance on another dataset without any adjustments. We employ two real-world datasets as testing sets, and the evaluation results are shown in Tab. \ref{tab:3}. It can be seen that our proposed GenePerson dataset outperforms all real-world and synthetic datasets, achieving mAP accuracies of 57.0\% and 56.1\% on Market-1501 and DukeMTMC-reID, respectively. Although our fast-generated GenePerson is only one-fifteenth the size of FineGPR, its remarkable improvements on real-world benchmarks demonstrate its effectiveness in modeling different identities and scene variations with finite training data. We attribute this superior result to the GenePerson with more diverse pedestrians and scenes. In particular, the best results are achieved when training on GenePerson using our PDM method, with further improvements in Rank-1 accuracy of 0.7\% and 1.4\% on the two real-world datasets, respectively. Our results emphasize that introducing disentangled text representations of style and content during training encourages the model to capture domain-invariant visual features.

\begin{table}[t]
	\centering
	\caption{Ablation study of the effectiveness of each loss component in PDM on Market-1501.}
	\normalsize
	\renewcommand\arraystretch{1.2}
	\setlength{\tabcolsep}{0pt}
	\begin{threeparttable} 
		\begin{tabular*}{\columnwidth}{@{\extracolsep{\fill}}>{\centering\arraybackslash}p{0.2cm}
				>{\centering\arraybackslash}p{1.2cm}
				>{\centering\arraybackslash}p{1.2cm}
				>{\centering\arraybackslash}p{1.2cm}
				>{\centering\arraybackslash}p{1.2cm}
				>{\centering\arraybackslash}p{1.2cm}|
				>{\centering\arraybackslash}p{1.1cm}
				>{\centering\arraybackslash}p{1.1cm}}
			\hline
%			\multicolumn{6}{c|}{Loss Components} & \multicolumn{2}{c}{Market-1501} \\
%			\hline
			& $\mathcal{L}_{\text{SupCon}}^1$ & $\mathcal{L}_{\text{Re-ID}}^1$ & $\mathcal{L}_{\text{Tri-txt}}$ & $\mathcal{L}_{\text{Re-ID}}^2 $ & $\mathcal{L}_{\text{SupCon}}^2$ & Rank-1 & mAP \\
			\hline\hline
			a) & $\checkmark$ & - & - & - & - & 23.0 & 10.6 \\
			b) & $\checkmark$ & $\checkmark$ & - & - & - & 55.0 & 29.9 \\
			c) & $\checkmark$ & $\checkmark$ & $\checkmark$ & - & - & 56.4 & 31.7 \\
			d) & $\checkmark$ & $\checkmark$ & $\checkmark$ & $\checkmark$ & - & 57.3 & 32.4 \\
			e) & $\checkmark$ & $\checkmark$ & $\checkmark$ & $\checkmark$ & $\checkmark$ & \textbf{57.7} & \textbf{32.6} \\
			\hline
		\end{tabular*}
	\end{threeparttable}
	\label{tab:4}
\end{table}

\begin{table}[t]
	\centering
	\caption{Ablation of training with different prompt strategies, including generic prompt, content prompt, and style-content prompt, evaluated on Market-1501.}
	\normalsize
	\renewcommand\arraystretch{1.2}
	\setlength{\tabcolsep}{1pt}
	\begin{threeparttable} 
		\begin{tabular*}{\columnwidth}{@{\extracolsep{\fill}}>{\centering\arraybackslash}p{4.1cm}|>{\centering\arraybackslash}p{1.5cm}>{\centering\arraybackslash}p{1.5cm}>{\centering\arraybackslash}p{1.5cm}}
			\hline
%			\multirow{2}{*}{ Method } & \multicolumn{3}{c}{ Market-1501 } \\
%			\cline{2-4}
			Method & Rank-1 & Rank-5 & mAP \\
			\hline\hline generic prompt  & 54.2 & 70.7 & 29.0 \\
			content prompt & 55.8 & 71.2 & 31.5 \\
			style-content prompt & \textbf{56.4} & \textbf{71.3} & \textbf{31.7} \\
			\hline
		\end{tabular*}	
	\end{threeparttable}
	\label{tab:5}
\end{table}

\begin{table}[t]
	\centering
	\caption{Effectiveness of the Global Feature-based Local Filtering Module and Sensitivity Analysis to the Number of Local Features $k$ on Market-1501.}
	\normalsize
	\renewcommand\arraystretch{1.2}
	\setlength{\tabcolsep}{1pt}
	\begin{threeparttable} 
		\begin{tabular*}{\columnwidth}{@{\extracolsep{\fill}}>{\centering\arraybackslash}p{2.75cm}|>{\centering\arraybackslash}p{1.5cm}|>{\centering\arraybackslash}p{1.25cm}>{\centering\arraybackslash}p{1.25cm}>{\centering\arraybackslash}p{1.25cm}}
			\hline
			Method & $k$ & mAP & Rank-1 & Rank-5 \\
			\hline\hline
			w/o filter module & 24 & 30.6 & 54.5 & 70.2 \\
			\hline
			\multirow{5}{*}{w/ filter module}
			& 4  & 31.7 & 55.9 & 71.7\\
			& 6 (Ours) & \textbf{32.6} & \textbf{57.7} & \textbf{78.7} \\
			& 8  & 32.0 & 56.6 & 72.1 \\
			& 10 & 31.9 & 55.9 & 72.1 \\
			& 12 & 30.9 & 55.3 & 70.9 \\
			\hline
		\end{tabular*}
		
	\end{threeparttable}
	\label{tab:6}
\end{table}

\begin{figure}[t]
	\centering
	\includegraphics[width=1.0\columnwidth]{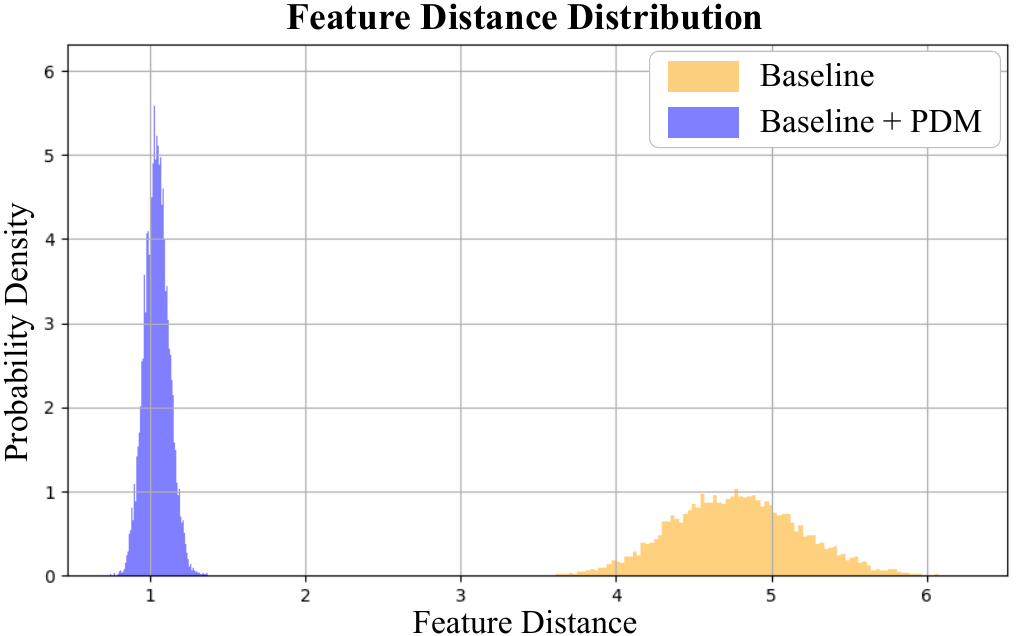}
	\caption{Visualization of distance distributions between randomly selected cross-domain sample pairs from Market-1501 and GenePerson, before and after applying PDM.}
	\label{fig:6}
\end{figure}

\subsection{Ablation Study}
%Below we perform ablation studies of the required elements of PDM. Notably, all experiments are trained on our GenePerson.
%% and tested on two real datasets, Market-1501 and MSMT17.
\textbf{Ablation Study on Loss Components.} In Tab. \ref{tab:4}, we evaluate the contribution of each loss component to model performance through incremental addition of loss terms. Row a) serves as the baseline, using only contrastive supervision $\mathcal{L}_{\text{SupCon}}^1$ on the initial pseudo-words. Comparing row b) with row a), we observe a significant improvement after introducing $\mathcal{L}_{\text{Re-ID}}^1$ to enforce identity consistency during pseudo-word learning, showing that identity supervision is crucial for model training. In row c), a triplet constraint $\mathcal{L}_{\text{Tri-txt}}$ is applied to the text features corresponding to the content pseudo-words, which contributes to enhancing discriminability in the semantic space. Furthermore, building on the joint optimization in the pseudo-word token learning module, the introduction of the prompt-driven style disentanglement module that employs $\mathcal{L}_{\text{Re-ID}}^2$ to learn image content leads to performance gains, as shown in row d). This confirms that the de-stylization projector can acquire domain-invariant content features for identity recognition. Finally, as seen in row e), the best performance is achieved when $\mathcal{L}_{\text{SupCon}}^2$ is added to align the visual representations with the content text representations.

\textbf{Ablation Study on Style-Content Prompt.} To explore whether the disentangled modeling strategy of style and content pseudo-word tokens provides a more granular guide for generalizable learning of visual features, we design three sets of comparison experiments in Tab. \ref{tab:5}. We train a baseline model that relies only on a generic template “a photo of a person” without any pseudo-word tokens to provide semantic guidance. We then use “a photo of a $[C^*]$ person” and “a $[S_1^*, \cdots, S_k^*]$ style of a $[C^*]$ person” as prompts in turn to explore the effects of style and content pseudo-words. The results in Tab. \ref{tab:5} show that the mAP improves by 2.5$\%$ when the content pseudo-word is incorporated into the template. When style and content are jointly modeled, the model achieves optimal performance with 31.7$\%$ mAP. This result emphasizes that modeling visual concepts at different semantic levels can effectively enhance the generalization of the model.

\textbf{Ablation Study on GFLFM.} We analyze the impact of the global feature-based local feature filtering strategy under different designs on the generalization performance. We disable the filtering strategy by setting $k=n$. The results in Tab. \ref{tab:6} show that introducing GFLFM outperforms the setting without the filtering strategy (w/o filter module) in various evaluation metrics, which validates the effectiveness of our strategy. In addition, to further determine the optimal number of effective local style features $k$, we vary $k$ from 4 to 12 for sensitivity analysis. It is observed that the model performance first improves and then decreases as $k$ increases, reaching peak performance when $k$ is taken as 6, with Rank1 and mAP accuracies of 57.7\% and 32.6\%, respectively. It is reasonable that an appropriate increase the number of potential style features provides richer information. However, too many local style features may introduce redundant visual information, leading to overfitting and increasing computational costs.

\textbf{Qualitative Visualization of PDM.} To visually understand and validate the effectiveness of our proposed PDM method, we conduct a qualitative analysis. Fig. \ref{fig:6} presents the distribution of distances between 10,000 randomly selected pairs of cross-domain samples from the real and virtual domains, before and after applying PDM. The results indicate that the distances between cross-domain samples are significantly reduced and more centralized after disentanglement, which confirms our hypothesis about prompt-driven style disentanglement. This indicates that PDM successfully weakens the effect of domain style differences and improves feature consistency, helping the model learn more domain-invariant content representations.

\section{Conclusion}
In this paper, we propose an innovative virtual dataset construction pipeline that employs a text-to-image diffusion model to directly synthesize virtual samples using generated prompts that incorporate multiple pedestrian attributes. This pipeline not only streamlines the data construction process but also improves the diversity and quality of generated data. In addition, we propose a simple yet effective generalization strategy, PDM. PDM utilizes the aligned multimodal potential space provided by CLIP, which mitigates the impact of style differences between virtual and real-world images by guiding the model to focus on domain-invariant content information through prompts. Extensive experiments validate the superiority of GenePerson and the effectiveness of PDM. Looking ahead, we will extend our dataset and method to handle more challenging tasks such as pose estimation, body part segmentation, and cross-modal retrieval.

\bibliographystyle{IEEEtran}
\bibliography{reference}

\end{document}